\documentclass{article}
\usepackage[accepted]{eiml_icml2026_local}
\usepackage{microtype}
\usepackage{graphicx}
\usepackage{booktabs}
\usepackage{hyperref}
\usepackage{amsmath}
\usepackage{amssymb}
\usepackage{xspace}
\usepackage{multirow}

\newcommand{\thinkH}{\ensuremath{\mathcal{H}_{\text{think}}}\xspace}
\newcommand{\ansH}{\ensuremath{\mathcal{H}_{\text{ans}}}\xspace}

\icmltitlerunning{When Thinking Hurts: Epistemic Signals in Reasoning-Chain VLMs}

\begin{document}

\twocolumn[
\icmltitle{When Thinking Hurts: Epistemic Signals in the\\
Reasoning Chains of Visual Language Models}

\icmlsetsymbol{equal}{*}

\begin{icmlauthorlist}
\icmlauthor{Mayank Singal}{anon}
\end{icmlauthorlist}
\icmlaffiliation{anon}{Independent Researcher}
\icmlcorrespondingauthor{Mayank Singal}{mayanksingal7@gmail.com}

\vskip 0.3in
]

\printAffiliationsAndNotice{}

\begin{abstract}
Uncertainty quantification for visual language models (VLMs) conventionally
targets the answer token distribution.  We provide the first three-family
empirical characterisation of answer entropy behaviour in thinking-mode VLMs.
Running four models on identical POPE adversarial samples, we find three
qualitatively distinct patterns: Qwen3-VL-8B-Thinking shows complete
collapse (ans\,H AUROC\,=\,0.492); GLM-4.1V-9B-Thinking shows no collapse
(0.716); and InternVL3-8B shows selective thinking (chains on only 50\%
of samples, ans\,H\,=\,0.675 full / 0.602 thinking-only).  Across all
three thinking-mode models, thinking chain entropy outperforms answer
entropy on the subset where chains are generated (0.647, 0.759, 0.608
vs.\ 0.492, 0.716, 0.602 respectively), suggesting chain signals are the
more reliable predictor whenever chains are present. This holds strongly for Qwen
and GLM, but with only marginal and statistically unreliable advantage for
InternVL3 ($n_{\text{FP}}$\,=\,17).  A 300-sample VQAv2
pilot confirms chain entropy (0.680) outperforms answer entropy (0.595)
on VQAv2 questions, with the gap largest for free-form answers
(0.733 vs.\ 0.467).  On harder reasoning
tasks (HallusionBench) both Qwen models show moderate signal
($\approx$\,0.64), consistent with incomplete pre-commitment on difficult
questions.  We additionally document structured abstention affecting
12--22\% of queries with asymmetry toward absent-object queries, and a
practical abstention gate raising accuracy from 71.0\% to 93.8\% at
62.7\% coverage with no additional inference cost.

\end{abstract}

\section{Introduction}
\label{sec:intro}

A growing class of VLMs generates an explicit
\texttt{<think>}\ldots\texttt{</think>} reasoning chain before committing to
an answer.  These thinking-mode models represent a qualitative shift in how
VLMs operate: rather than mapping inputs directly to output distributions,
they first conduct extended deliberation whose result then conditions the
answer.  This architectural change has a direct consequence for uncertainty
quantification that has not previously been characterised: the thinking chain
acts as a strong conditioning prior that collapses the answer token
distribution, destroying the epistemic signal that conventional uncertainty
methods depend on.

We demonstrate this collapse directly through a controlled ablation.
Running Qwen3-VL-8B-Thinking and Qwen3-VL-8B-Instruct (the same model
family, same vision encoder, differing only in whether a thinking chain is
generated) on identical POPE adversarial samples, answer entropy achieves
AUROC\,=\,0.899 for hallucination detection in the non-thinking model but
falls to AUROC\,=\,0.492 (chance) in the thinking model.  This is not a
task-structural artefact: the questions, images, and evaluation protocol are
identical.  The thinking mode itself is responsible.

This failure creates a practical problem: thinking-mode VLMs are
increasingly deployed in settings where reliable uncertainty estimates matter,
yet every existing output-level uncertainty method produces no useful signal
for this model class under greedy decoding.  We show that the thinking chain
itself restores the signal, providing two robust hallucination predictors
extractable from a single forward pass at no additional inference cost.

We make four contributions:

\begin{enumerate}
\item We provide the first three-family empirical characterisation of answer
  entropy behaviour in thinking-mode VLMs, revealing three qualitatively
  distinct patterns: complete collapse (Qwen3-VL-8B-Thinking, 0.492),
  no collapse (GLM-4.1V-9B-Thinking, 0.716), and selective thinking
  (InternVL3-8B, 50\% chain rate, 0.675 full / 0.602 thinking-only).
  These patterns demonstrate that collapse is training-recipe-dependent,
  not a universal property of thinking-mode decoding.

\item We show that thinking chain signals are \emph{consistently superior}
  to answer entropy across the two families with sufficient statistical power (Qwen and GLM):
  \thinkH achieves 0.647 (Qwen) and 0.759 (GLM) vs.\ answer entropy of
  0.492 and 0.716 respectively.  Chain signals are the recommended approach
  for any thinking-mode VLM, whether or not collapse has occurred, though
  generalisation beyond these two families requires further verification.

\item We document \emph{structured abstention} that replicates across
  benchmarks with consistent asymmetry, but through mechanistically distinct
  failure modes: over-deliberation on POPE vs.\ immediate chain collapse on
  HallusionBench.

\item We demonstrate a practical abstention gate requiring no additional
  inference: refusing unclear samples and long-chain committed predictions
  raises accuracy from 71.0\% to 93.8\% at 62.7\% coverage on POPE.
\end{enumerate}

We conjecture these findings reflect structural properties of thinking-mode
decoding; verification across additional model families is ongoing.

\section{Background and Related Work}
\label{sec:related}

\paragraph{Uncertainty in VLMs.}
VL-Uncertainty~\citep{zhang2024vluncertainty} applies semantic
entropy~\citep{farquhar2024detecting} to VLM outputs under prompt
perturbation, requiring multiple forward passes. UMPIRE~\citep{lau2026umpire}
computes an incoherence-adjusted semantic volume over sampled MLLM responses,
treating each response atomically; applied to thinking-mode VLMs it would
conflate the pre-commitment reasoning phase with the committed answer, and
cannot detect answer entropy collapse.  HALP~\citep{kogilathota2026halp}
probes frozen representations before generation, requiring a trained probe
and an extra forward pass.  Our approach requires neither.

\paragraph{Thinking-mode VLMs.}
Extended reasoning can worsen visual grounding~\citep{thierry2025morethinking}
and CoT has been shown to induce overconfidence at the answer-token
level~\citep{wang2026cost}, which we confirm and extend here with a controlled
ablation.  \citet{wang2025knowinganswer} show that VLMs can produce correct
answers via flawed reasoning paths; we demonstrate the converse.
\citet{singh2025thinkright} study chain length for accuracy; we show it
simultaneously carries epistemic information.

\paragraph{Self-consistency and self-knowledge.}
Self-consistency~\citep{wang2022selfconsistency} samples diverse reasoning
chains and marginalises to the most common answer, using reasoning diversity
for accuracy rather than as an uncertainty signal, and requires multiple
passes.  Our approach uses a single greedy chain.
\citet{kadavath2022know} show that LLMs exhibit measurable self-knowledge
about their own correctness; our chain signals can be viewed as a
multimodal extension of this line, where the deliberation process itself
encodes uncertainty rather than a post-hoc verbalized confidence.

\section{Setup}
\label{sec:setup}

\paragraph{Models.}
\textbf{Qwen3-VL-8B-Thinking}~\citep{qwen2025vl}: generates a
\texttt{<think>}\ldots\texttt{</think>} block before answering under greedy
decoding (\texttt{do\_sample=False}).  \textbf{Qwen3-VL-8B-Instruct}: the
same model family without thinking-mode generation, used as a non-thinking
control.  Both run in \texttt{bfloat16} on a single A100-SXM4-40\,GB GPU.

\paragraph{Benchmarks.}
\textbf{POPE adversarial}~\citep{li2023pope}: 1{,}000 binary yes/no questions
about object presence in COCO val2014 images, balanced labels.
\textbf{HallusionBench}~\citep{guan2024hallusion}: 951 visual questions
(178 text-only excluded as they engage no visual grounding pathway) covering
visual illusions, edited images, charts, and visual reasoning; imbalanced
labels (550 no / 401 yes).

\paragraph{Signals.}
Per-token Shannon entropy within the \texttt{<think>} block:
$H_t = -\sum_{v} p_t(v)\log p_t(v)$.
We record \thinkH\,=\,mean of $H_t$ over thinking tokens; \ansH\,=\,mean
entropy over answer tokens; $L$\,=\,number of thinking tokens.  All are free
byproducts of a single greedy forward pass.

\section{Results}
\label{sec:results}

\subsection{Thinking Mode Causes Answer Entropy Collapse}
\label{sec:ablation}

Table~\ref{tab:ablation} presents the core ablation: Qwen3-VL-8B-Thinking
vs.\ Qwen3-VL-8B-Instruct on the same 1{,}000 POPE adversarial samples.

\begin{table}[h]
\centering
\caption{Four-model comparison on POPE adversarial (1,000 samples each).
  \ansH and \thinkH are conditioned AUROCs (FP vs.\ TP within yes-predictions).
  $^\dagger$Full sample. $^\ddagger$Thinking-only subset ($n_{\text{yes}}$\,=\,284).}
\label{tab:ablation}
\smallskip
\setlength{\tabcolsep}{4pt}
\begin{tabular}{lcccccc}
\toprule
Model & Acc & Unk & Thk & FP & \ansH & \thinkH \\
\midrule
Qwen-Instruct  & 87.4 & 0\%  & 0\%  & 51 & .899 & ---  \\
Qwen-Thinking  & 71.0 & 22\% & 100\%& 34 & .492 & .647 \\
GLM-Thinking   & 86.4 &  1\% & 99\% & 35 & .716 & \textbf{.759} \\
InternVL3      & 87.2 &  1\% & 50\% & 40 & .675$^\dagger$ & .608$^\ddagger$ \\
\bottomrule
\end{tabular}
\end{table}

Table~\ref{tab:ablation} presents the four-model cross-family comparison
on 1{,}000 POPE adversarial samples, revealing three qualitatively distinct
patterns.

\textit{Complete collapse} (Qwen3-VL-8B-Thinking): answer entropy AUROC
collapses to 0.492 (chance), while thinking chain entropy achieves 0.647.
The model generates thinking chains on every sample but fully pre-commits
before the answer token, destroying answer entropy variance.

\textit{No collapse} (GLM-4.1V-9B-Thinking): answer entropy AUROC remains
0.716 despite explicit thinking chains, and chain entropy (0.759) still
outperforms it.  GLM's recipe does not achieve full pre-commitment.

\textit{Selective thinking} (InternVL3-8B): chains are generated on only
50\% of samples.  On the full sample, answer entropy achieves 0.675; on the
thinking-only subset ($n$\,=\,284 yes-predictions, $n_{\text{FP}}$\,=\,17),
chain entropy (0.608) marginally outperforms answer entropy (0.602).
This margin is small and the FP count is low; we do not claim a
statistically reliable advantage for InternVL3's chain content.
The more robust signal for this model is chain \emph{presence}: the FP rate
is more than twice as high when InternVL3 skips thinking (13.1\%) compared
to when it generates a chain (6.0\%).  For InternVL3, chain absence carries
much of the epistemic signal, while chain content adds comparatively little.

Across all three thinking-mode models, the chain-level signal is competitive
with or superior to answer entropy on samples where chains are generated:
strongly so for Qwen (0.647 vs.\ 0.492) and GLM (0.759 vs.\ 0.716),
marginally so for InternVL3 (0.608 vs.\ 0.602 on the thinking-only subset).
Chain presence or absence is a reliable signal for InternVL3 regardless of
chain content.

\begin{figure}[h]
\centering
\includegraphics[width=\columnwidth]{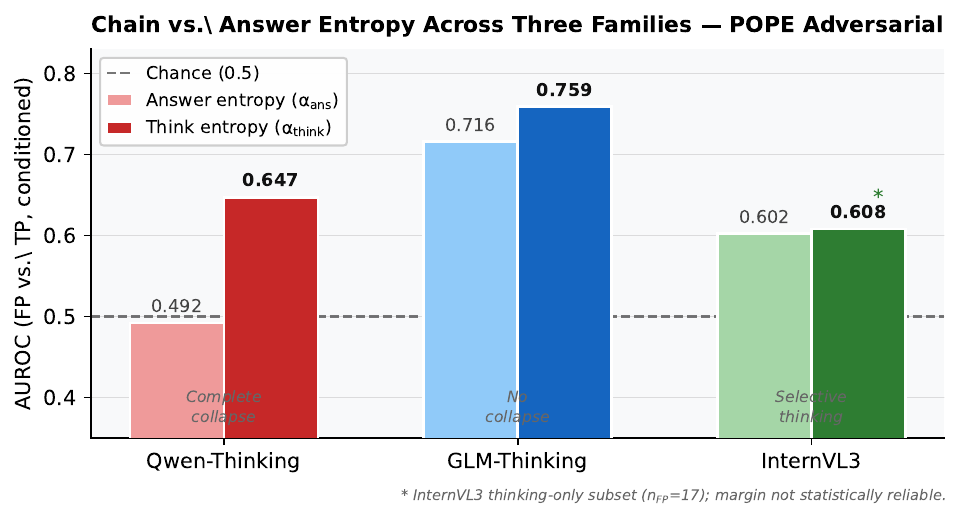}
\caption{Chain vs.\ answer entropy AUROC across three thinking-mode model
  families on POPE adversarial.  Three qualitatively distinct patterns:
  complete collapse (Qwen), no collapse (GLM), and selective thinking
  (InternVL3).  Chain entropy (dark bars) is competitive with or superior
  to answer entropy (light bars) in all cases.}
\label{fig:crossfamily}
\end{figure}

\subsection{Answer Entropy Is Structurally Confounded}
\label{sec:ansent}

Table~\ref{tab:auroc_unconditional} reports unconditional AUROCs for all
signals on binary-committing samples from both benchmarks.

\begin{table}[h]
\centering
\caption{Unconditional AUROCs on binary-committing samples. For the thinking
  model, answer entropy direction is benchmark-dependent; thinking chain
  signals are consistently positive.}
\label{tab:auroc_unconditional}
\smallskip
\begin{tabular}{lcccc}
\toprule
 & \multicolumn{2}{c}{POPE ($n$\,=\,779)} &
   \multicolumn{2}{c}{HalluBench ($n$\,=\,831)} \\
\cmidrule(lr){2-3}\cmidrule(lr){4-5}
Signal & FP & Error & FP & Error \\
\midrule
\ansH         & 0.392 & 0.554 & 0.726 & 0.602 \\
\thinkH       & 0.526 & 0.636 & 0.768 & 0.768 \\
Chain len.\ $L$ & 0.573 & 0.686 & 0.650 & 0.630 \\
\bottomrule
\end{tabular}
\end{table}

On POPE, \ansH\,=\,0.392 for FP prediction (below chance), meaning higher
answer confidence predicts \emph{more} hallucination.  This inversion arises
from the answer-direction confound: the model assigns higher confidence to
positive predictions than negative ones regardless of correctness.  On
HallusionBench, \ansH is positively predictive (0.726) because the harder
questions create genuine answer distribution variance not dominated by this
confound.  Thinking chain signals are consistently positive on both.

\subsection{Conditioning on Answer Direction Recovers Strong Signal}
\label{sec:conditioned}

Restricting to yes-predictions eliminates the answer-direction confound.
Table~\ref{tab:auroc_conditioned} reports per-benchmark and pooled results.

\begin{table}[h]
\centering
\caption{Conditioned AUROCs (FP vs.\ TP within yes-predictions;
  $n_{\text{FP}}^{\text{POPE}}$\,=\,34, $n_{\text{FP}}^{\text{HB}}$\,=\,129,
  pooled $n_{\text{FP}}$\,=\,163).
  Bootstrap 95\% CIs (10{,}000 resamples); $|r|$ from Mann-Whitney $U$.}
\label{tab:auroc_conditioned}
\smallskip
\setlength{\tabcolsep}{4pt}
\begin{tabular}{lccccc}
\toprule
Signal & POPE & HB & Pooled & 95\% CI & $|r|$ \\
\midrule
\ansH       & .492 & .662 & .432 & [.386,\,.478] & ---  \\
\thinkH     & .647 & .764 & .786 & [.745,\,.825] & .53  \\
Chain len.\ & .710 & .708 & .776 & [.735,\,.815] & .42  \\
\bottomrule
\end{tabular}
\end{table}

The pooled answer entropy CI\,=\,[0.386,\,0.478] lies entirely below chance
across 163 FPs and 1{,}951 samples. Pooled thinking entropy achieves
AUROC\,=\,0.786 (lower CI\,=\,0.745); chain length achieves 0.776
(lower\,=\,0.735). Both are significant at $p<0.001$.

Pooling across benchmarks is appropriate for the aggregate comparison:
thinking chain signals are consistently superior to answer entropy on each
benchmark individually, and the pooled estimate characterises this advantage
across a broader task distribution, even as the dominant signal within that
family varies by task structure. Chain length dominates on POPE (0.710
vs.\ 0.647), where hallucinations are committed quickly.  Thinking entropy
dominates on HallusionBench (0.764 vs.\ 0.708; pooled $|r|$\,=\,0.53, Table~\ref{tab:auroc_conditioned}), where FP
chains are 83\% longer (322.6 vs.\ 176.2 tokens) and 47\% more uncertain
than TP chains. When a model works hard and \emph{still} commits
incorrectly, entropy captures that struggle better than length alone.

Conditioning on answer direction is a diagnostic for understanding the
confound mechanism, not a deployment prescription.  In deployment, the
ground-truth label is unavailable; practitioners use unconditional signals.
Table~\ref{tab:auroc_unconditional} shows chain length achieves
AUROC\,=\,0.686 for error prediction \emph{unconditionally}; the signal is
useful without conditioning, and conditioning reveals why it works.

\subsection{Epistemic Hierarchy}
\label{sec:hierarchy}

Both chain length and thinking entropy increase monotonically from
committed-correct (TP: $L$\,=\,59.9, \thinkH\,=\,0.218 on POPE) through
committed-wrong (FP: 74.4, 0.262) to abstaining-with-chain
(124.3, 0.338), with the same ordering on HallusionBench at different
absolute scales.  This monotonic hierarchy confirms that chain signals
track epistemic state rather than surface-level output properties.

\subsection{Structured Abstention}
\label{sec:abstention}

Abstention rates are 22.1\% on POPE and 12.6\% on HallusionBench, with
consistent asymmetry toward absent-object queries (31.4\% vs.\ 12.8\%
label-no vs.\ label-yes on POPE; 14.2\% vs.\ 10.5\% on HallusionBench).
Mechanisms differ: on POPE, 199/221 abstentions involve extended chains
(mean 124.3 tokens) that fail to resolve; on HallusionBench, 118 of the
120 abstentions produce no chain at all even at 1{,}024 tokens, and 119
remain unresolved after rerunning the 231 originally-unclear samples at
extended token budget.

\subsection{Abstention Gate}
\label{sec:gate}

\begin{figure}[h]
\centering
\includegraphics[width=\columnwidth]{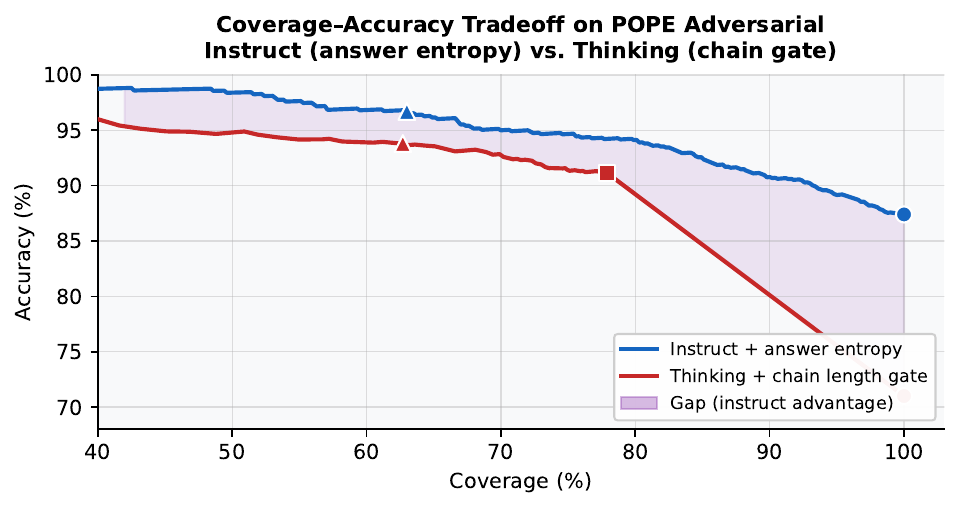}
\caption{Coverage--accuracy tradeoff on POPE adversarial.  \emph{Note}: the
  instruct model (blue) is not a deployment alternative when a thinking-mode
  VLM is already in use; its curve is a reference ceiling only.  For
  thinking-mode deployment, answer entropy provides no signal
  (AUROC\,=\,0.492, chance); the chain gate (red) is the only available
  option, recovering accuracy from 71.0\% to 93.8\% at 62.7\% coverage.
  The gap narrows from 16\,pp at 100\% coverage to $\approx$3\,pp at 40\%.}
\label{fig:coverage}
\end{figure}

Table~\ref{tab:gate} shows the accuracy-coverage tradeoff for both
benchmarks, and compares the thinking model's chain gate against the
instruct model's answer entropy gate on POPE.

\begin{table}[h]
\centering
\caption{Accuracy-coverage tradeoff. ``Top $X$\%'' thresholds are computed
  over committed samples only; coverage is reported over all 1{,}000 samples.
  On POPE, the thinking model's chain gate (93.8\% at 62.7\%) is compared
  against the instruct model using answer entropy (96.7\% at 63.0\%).}
\label{tab:gate}
\smallskip
\begin{tabular}{lrr}
\toprule
Gate & Coverage & Accuracy \\
\midrule
\multicolumn{3}{l}{\textit{POPE adversarial --- thinking model}} \\
\quad None (baseline)          & 100.0\% & 71.0\% \\
\quad Refuse unclear           &  77.9\% & 91.1\% \\
\quad +top 20\% chain refused  &  62.7\% & 93.8\% \\
\quad +top 30\% chain refused  &  54.9\% & 94.2\% \\
\addlinespace
\multicolumn{3}{l}{\textit{POPE adversarial --- instruct model (answer entropy)}} \\
\quad None (baseline)                & 100.0\% & 87.4\% \\
\quad +top 20\% entropy refused      &  80.0\% & 94.1\% \\
\quad +top 37\% entropy refused      &  63.0\% & 96.7\% \\
\addlinespace
\multicolumn{3}{l}{\textit{HallusionBench --- thinking model}} \\
\quad None (baseline)          & 100.0\% & 67.2\% \\
\quad Refuse unclear           &  87.4\% & 76.9\% \\
\quad +top 20\% chain refused  &  69.9\% & 80.2\% \\
\bottomrule
\end{tabular}
\end{table}

At 100\% coverage, the instruct model (87.4\%) substantially outperforms
the thinking model (71.0\%) on POPE, with the gap explained almost entirely
by the thinking model's 22.1\% abstention rate (committed accuracy: 91.1\%).
The chain gate progressively recovers accuracy, reaching 93.8\% at 62.7\%
coverage, but the instruct model using answer entropy achieves 96.7\% at
matched coverage.  The chain gate thus closes but does not eliminate the
practical gap introduced by thinking-mode abstention.

This comparison clarifies the role of chain signals in deployment. When a
thinking-mode VLM is used, answer entropy provides no useful gating signal
on easy binary tasks (AUROC\,=\,0.492 on POPE); the chain gate is the
\emph{only available} signal and substantially improves on the uncalibrated
baseline.  Whether the accuracy-coverage tradeoff justifies using a
thinking-mode VLM over an instruct model is an architectural decision beyond
this paper's scope; our contribution is identifying the signals that make
thinking-mode deployment epistemically tractable.

The full coverage-accuracy curves show the instruct model's answer entropy
gate dominates at all tested coverage levels, with the gap narrowing from
16 percentage points at 100\% coverage to approximately 3 points at 40\%
coverage; the chain gate substantially closes but does not eliminate the
gap.  The curve is smooth with no sharp elbow, indicating threshold
selection is not sensitive to a specific operating point; any threshold
in the 50--70\% coverage range yields comparable accuracy improvements.
Thinking entropy is more transferable across benchmarks than chain
length: the POPE top-20\% entropy threshold applied to HallusionBench
without recalibration achieves 91.0\% accuracy at 45.5\% coverage, versus
88.7\% at 23.2\% for chain length.  Practitioners requiring cross-task
transfer should prefer entropy thresholds.

\subsection{Generalisation to Open-Ended VQA}
\label{sec:vqa}

To test whether chain signals generalise beyond binary tasks, we run
Qwen3-VL-8B-Thinking on 300 VQAv2 samples stratified across answer types.

\begin{table}[h]
\centering
\caption{VQAv2 pilot (Qwen3-VL-8B-Thinking, $n$\,=\,300).  Error prediction
  AUROCs (wrong vs.\ correct).  Think entropy outperforms answer entropy
  across all answer types; the gap is largest for free-form answers where
  the binary confound is weakest.}
\label{tab:vqa}
\smallskip
\setlength{\tabcolsep}{4pt}
\begin{tabular}{lcccc}
\toprule
Answer type & $n$ & Acc & \ansH & \thinkH \\
\midrule
Yes/No      & 99  & 73\% & 0.668 & \textbf{0.739} \\
Number      & 77  & 65\% & 0.682 & 0.676 \\
Other (free-form) & 97 & 65\% & 0.467 & \textbf{0.733} \\
\midrule
All         & 273 & 68\% & 0.595 & \textbf{0.680} \\
\bottomrule
\end{tabular}
\end{table}

Chain entropy outperforms answer entropy on all answer types pooled
(0.680 vs.\ 0.595).  The free-form ``other'' subset is most revealing:
answer entropy falls below chance (0.467): the partial answer-direction
confound persists even without strict binary structure, while chain entropy
remains strongly predictive (0.733).  This suggests the chain signal
advantage is not an artefact of binary task structure.

\section{Discussion}
\label{sec:discussion}

\paragraph{Connection to the broader UQ literature.}
The collapse we document is an instance of a general phenomenon: conditioning
on additional information reduces posterior variance~\citep{mackay1992practical}.
In ensemble and MC-dropout frameworks, epistemic uncertainty is estimated from
variance across samples of $p(a \mid x)$; the thinking chain eliminates this
variance not through calibration but by pre-committing the answer.  Chain
entropy recovers a proxy for pre-commitment epistemic uncertainty, analogous
to the epistemic/aleatoric decomposition of~\citet{hullermeier2021aleatoric}:
chain entropy reflects uncertainty about the correct reasoning path;
answer entropy post-collapse reflects only aleatoric residuals (or zero).

\paragraph{Uncertainty-theoretic interpretation.}
The answer entropy collapse can be understood as a consequence of
conditioning.  If $z$ denotes the thinking chain and $a$ the answer,
$H(a \mid x, z)$ decreases as $z$ more fully resolves the question.
When the chain \emph{fully} pre-commits to an answer direction,
$H(a \mid x, z) \approx 0$, eliminating the epistemic variance in $p(a \mid x)$.
Our cross-family results show that the degree of collapse is
training-recipe-dependent: Qwen3-VL-8B-Thinking's recipe produces
near-complete pre-commitment on easy binary tasks (AUROC 0.492);
GLM-4.1V-9B-Thinking's recipe does not (AUROC 0.716), despite both models
generating explicit reasoning chains.  The chain-level signals---\thinkH
and $L$---are measured \emph{before} commitment completes and are robust
across both regimes: they outperform answer entropy for both thinking-mode
models regardless of whether collapse occurs.

\paragraph{Training recipes and open questions.}
GLM almost never abstains (1.3\% vs.\ 22.1\% for Qwen), suggesting its
recipe encourages committing under uncertainty rather than deliberating
to pre-commitment, preserving answer entropy variance at the cost of more
hallucinations on difficult questions.  Section~\ref{sec:vqa} shows chain
signals also outperform answer entropy on VQAv2 open-ended questions,
confirming the advantage is not confined to binary tasks.

\paragraph{Signal transferability.}
Chain length thresholds do not transfer across benchmarks because chain
length scales with question difficulty. Thinking entropy is more transferable
because it operates on a more benchmark-agnostic scale. Threshold selection
in deployment requires held-out calibration data from the target task
distribution.

\paragraph{Comparison with UMPIRE.}
UMPIRE~\citep{lau2026umpire} embeds the full output---thinking chain and
answer---as a single vector, unable to detect the answer entropy collapse we
document. It would report similar uncertainty estimates for thinking-mode and
non-thinking models, missing the structural difference our ablation reveals.

\paragraph{Limitations and open questions.}
\textit{Model scope}: We evaluate three thinking-mode VLMs from distinct
families: Qwen3-VL-8B (Alibaba), GLM-4.1V-9B (THUDM/Zhipu AI), and
InternVL3-8B (OpenGVLab/Shanghai AI Lab), plus a non-thinking Qwen control.
All are at 8--9B scale; larger thinking-mode VLMs (32B, 72B) remain untested.
\textit{Decoding}: All results use greedy decoding.  Whether the collapse
patterns persist under stochastic sampling (temperature\,$>$\,0) is an
open question: sampling prevents full pre-commitment at the answer token
and may rehabilitate answer entropy for Qwen-style models.  This is the
single most actionable remaining experiment.
\textit{InternVL3 depth}: InternVL3-8B is evaluated only on the headline
POPE comparison (Table~\ref{tab:ablation}); the full analysis pipeline
(conditioned AUROCs, epistemic hierarchy, abstention gate) covers only
Qwen and GLM.  The InternVL3 results are sufficient to characterise its
pattern but do not support per-model chain signal comparisons beyond the
headline numbers.
\textit{Signal correlation}: Thinking entropy and chain length are
moderately correlated (Pearson $r$\,=\,0.70 on POPE, 0.44 on HallusionBench)
but explain only 50\% and 19\% of shared variance---practitioners without
logit access can use chain length alone.
\textit{Failure modes}: 11/34 FPs on POPE and 14/129 on HallusionBench
produce chain entropy below the TP median (``confident hallucinations'':
fast, low-entropy chains that commit to a wrong answer), representing the
hard cases chain signals cannot catch.
\textit{Baselines}: No empirical comparison against sampling-based methods
(e.g.\ self-consistency); our contribution is the zero-cost single-pass
signal.
\textit{Epistemic vs.\ aleatoric}: Chain entropy may partly reflect
inherent image ambiguity rather than model ignorance; disentangling these
requires controlled image manipulations.
\textit{Scope}: Both primary benchmarks are binary yes/no tasks.
Section~\ref{sec:vqa} reports a 300-sample VQAv2 pilot showing chain
signals generalise to open-ended questions; larger-scale open-ended
evaluation remains future work.

\section{Conclusion}
\label{sec:conclusion}

We provide the first controlled evidence that thinking-mode decoding causes
answer entropy collapse: answer entropy AUROC drops from 0.899 to 0.492 when
switching from Qwen3-VL-8B-Instruct to Qwen3-VL-8B-Thinking on identical
data.  For thinking-mode VLMs, the thinking chain itself restores
discriminative power: pooled thinking entropy (AUROC\,=\,0.786) and chain
length (AUROC\,=\,0.776) are robust hallucination predictors with tight
confidence intervals, while pooled answer entropy lies entirely below chance.
Structured abstention affecting 12--22\% of queries replicates across
benchmarks with consistent directional asymmetry but mechanistically distinct
failure modes.  These findings argue that uncertainty estimation for
thinking-mode VLMs must track the reasoning process rather than the committed
answer, and that the answer-direction confound should be routinely checked in
any thinking-chain uncertainty analysis.

\bibliography{main}
\bibliographystyle{abbrvnat}

\appendix

\section{Reproducibility}
\label{app:repro}
Qwen3-VL-8B-Thinking and Qwen3-VL-8B-Instruct in \texttt{bfloat16} on a
single A100-SXM4-40\,GB GPU. POPE adversarial and COCO val2014 are publicly
available. HallusionBench via \texttt{lmms-lab/HallusionBench} on
HuggingFace. Per-token entropy from the full softmax distribution over the
model vocabulary. Bootstrap CIs: 10{,}000 resamples,
\texttt{numpy.random.default\_rng(42)}. HallusionBench samples unclear at
512 tokens were rerun at 1{,}024 tokens; 112/231 resolved. All results use
a corrected six-group classification (TP, TN, FP, FN, unclear\_chain,
unclear\_nochain). Code released upon acceptance.

\section{Cross-Family Control Details}
\label{app:control}
\paragraph{Qwen3-VL-8B-Instruct.}
Produces no thinking tokens on any of the 1{,}000 POPE adversarial or 951
HallusionBench samples.  Answer entropy achieves AUROC\,=\,0.899 on POPE
and 0.627 on HallusionBench, confirming the signal is intact for
non-thinking models.

\paragraph{GLM-4.1V-9B-Thinking (THUDM).}
Uses \texttt{Glm4vForConditionalGeneration} architecture (distinct from
Qwen3VL), \texttt{</think>} token ID 151346, and wraps answers in
\texttt{<answer><|begin\_of\_box|>}\allowbreak\texttt{yes<|end\_of\_box|></answer>} format.
On 1{,}000 POPE adversarial samples: accuracy 86.4\%, FP count 35,
unclear rate 1.3\%, \ansH AUROC 0.716, \thinkH AUROC 0.759.
Answer entropy does not collapse for GLM; chain entropy still outperforms it.

\paragraph{InternVL3-8B (OpenGVLab/Shanghai AI Lab).}
Uses \texttt{InternVLForConditionalGeneration} architecture, \texttt{</think>}
as a 3-token sequence, activated via R1-style system prompt.  On 1{,}000
POPE adversarial samples: accuracy 87.2\%, FP count 40, unclear rate 1.2\%,
thinking triggered on 49.8\% of samples.  \ansH AUROC\,=\,0.675 (full
sample); on the thinking-only subset ($n_{\text{yes}}$\,=\,284,
$n_{\text{FP}}$\,=\,17): \ansH\,=\,0.602, \thinkH\,=\,0.608.  FP rate is
higher when thinking is absent (13.1\%) than present (6.0\%), suggesting
chain absence is itself an uncertainty signal for this model.

\section{Usage of Generative AI Tools}
The authors made use of AI-based writing tools during the manuscript
preparation process to assist with language refinement and editorial polish.

\end{document}